\documentclass[letterpaper,10pt,conference]{ieeeconf}

\IEEEoverridecommandlockouts
\usepackage{amsmath}
\usepackage{amssymb}
\usepackage[misc]{ifsym}
\usepackage{graphicx}
\usepackage{float}
\usepackage{placeins}
\usepackage{booktabs}
\usepackage{tabularx}
\usepackage{multirow}
\usepackage{makecell}
\usepackage{colortbl}
\usepackage{tikz}
\usetikzlibrary{arrows.meta,positioning,fit,calc}
\usepackage{xcolor}

\definecolor{c1}{HTML}{D6ECF0}
\definecolor{metablue}{HTML}{0064E0}
\definecolor{fgblue}{HTML}{DCEBFA}
\definecolor{mucosagreen}{HTML}{E1F1DD}
\definecolor{fovorange}{HTML}{F8E5C0}
\definecolor{invalidgray}{HTML}{E7E7E7}

\makeatletter
\let\NAT@parse\undefined
\makeatother

\usepackage[breaklinks,colorlinks]{hyperref}
\usepackage[capitalize]{cleveref}
\crefname{section}{Sec.}{Secs.}
\crefname{table}{Tab.}{Tabs.}
\crefname{figure}{Fig.}{Figs.}

\newcommand{\method}{\textsc{LAMP}}

\newcommand{\minisection}[1]{\vspace{0.04in}\noindent{\bf #1}\,}
\newcommand{\circlednum}[1]{%
    \tikz[baseline=(X.base)]{\node[draw,circle,inner sep=.3pt] (X) {#1};}%
}

\title{\bf Beyond the Foreground: FOV-Aware Polyp Image Synthesis via Lesion-Guided Adaptive Mucosal Context Propagation}

\author{Tong Wang$^{1,2}$, Yuting He$^{3}$, Bin Ren$^{2}$, Yutong Xie$^{2}$, Guanyu Yang$^{1}$\\
{\small $^{1}$Southeast University \quad $^{2}$MBZUAI \quad $^{3}$Case Western Reserve University}
}

\begin{document}

\maketitle
\thispagestyle{empty}
\pagestyle{empty}

\begin{abstract}
Synthetic image and mask pairs can alleviate scarce colonoscopy annotations, but realistic synthesis requires preserving the supplied lesion while generating compatible mucosa. Existing foreground-guided methods treat all non-foreground pixels as background and rely mainly on local integration. Directly applying them to colonoscopy causes two problems: non-mucosal black regions contaminate generated tissue, and local reasoning produces inconsistent mucosal texture and illumination. We propose LAMP, the first foreground-guided framework for polyp image synthesis based on lesion-guided adaptive mucosal context propagation. LAMP explicitly separates the lesion, valid mucosa, and camera exterior using a field-of-view (FOV) mask. Lesion-to-Mucosa cross-attention extracts lesion appearance conditions for valid-mucosa locations, while FOV-constrained multidirectional Vision Receptance Weighted Key Value propagates them over legal tissue support. An adaptive gate then controls their residual fusion into the diffusion U-Net. Extensive experiments on five polyp datasets demonstrate that LAMP substantially outperforms existing methods in overall generation quality and consistently improves five downstream segmentation models. Our code will be released at \url{https://github.com/wangtong627/LAMP}.
\end{abstract}

\section{Introduction}

Reliable polyp segmentation is a core perception capability for robot-assisted colonoscopy, enabling automated lesion detection, localization, measurement, and intervention planning~\cite{fan2020pranet}. Robust performance requires diverse, pixel-accurate training data, yet variations in lesion appearance, mucosal texture, illumination, and endoscopic devices are costly to collect and annotate. Synthetic polyp generation offers scalable augmentation for this perception pipeline, provided images are realistic and consistent with their masks.

Existing polyp generation methods commonly render an image from a mask, boundary, or candidate region. Polyp-DDPM~\cite{dorjsembe2024polypddpm}, ControlPolypNet~\cite{sharma2024controlpolypnet}, Polyp-Gen~\cite{liu2025polypgen}, MaskFactory~\cite{qian2024maskfactory}, and Polyp-LDM~\cite{qiu2025polypldm} improve structural control and visual quality in different ways. Nevertheless, the lesion RGB is still redrawn by a generator. The resulting appearance may cross, underfill, or blur the prescribed contour, weakening the correspondence required for segmentation supervision.
Foreground-guided generation provides a complementary formulation: retain a concrete, annotated foreground condition and generate a compatible environment around it. Classical Poisson compositing~\cite{dimartino2016poisson} requires a selected target canvas, while camouflage synthesis~\cite{chu2010camouflage} jointly designs object and context. LAKE-RED~\cite{zhao2024lakered} retrieves latent background knowledge from foreground features, FACIG~\cite{chen2025foreground} strengthens foreground-aware feature integration and reconstruction, and CamoDreamer~\cite{wang2026camodreamer} separates object and background rendering before contextual blending. These methods demonstrate that foreground appearance can organize environment generation without a manually selected background.

\begin{figure}
    \centering
    \includegraphics[width=1\linewidth]{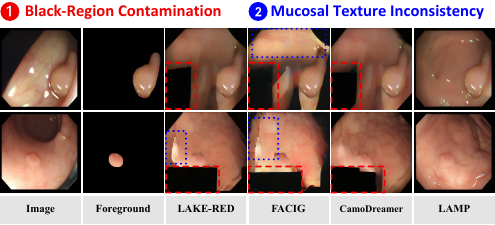}
    \vspace{-6.5mm}
    \caption{\textbf{Challenges in polyp image synthesis.}  (1) \emph{Black-region contamination}: non-mucosal black regions leak into synthesized tissue. (2) \emph{Mucosal texture inconsistency}: generated mucosa contains incoherent textures.
    Columns 1 and 2 show the original image and foreground, respectively. Columns 3 to 6 show polyp images generated from the foreground by LAKE-RED, FACIG, CamoDreamer, and LAMP (Ours).
    In compared methods, red and blue boxes mark the first and second challenge, respectively.}
    \label{fig:motivation}
    \vspace{-6.5mm}
\end{figure}

Directly transferring this formulation to colonoscopy, however, leaves two domain-specific challenges unresolved. (1) \emph{Black-Region Contamination.} Besides the supplied lesion and valid mucosa, endoscopic images may contain non-mucosal black regions introduced by field-of-view (FOV) boundaries, cropping, or privacy masking. Treating every non-foreground pixel as background allows these invalid regions to contaminate mucosal generation. (2) \emph{Mucosal Texture Inconsistency.} Local foreground integration does not explicitly propagate lesion-dependent appearance across the mucosal field, often producing blurry textures, uneven illumination, and discontinuous folds or vessels. As highlighted in \cref{fig:motivation}, LAKE-RED, FACIG, and CamoDreamer exhibit both limitations in their synthesized colonoscopy images.

To address these challenges, we propose \method, short for \textbf{L}esion-Guided \textbf{A}daptive \textbf{M}ucosal Context \textbf{P}ropagation. Given a polyp foreground, its mask, and an endoscopic FOV mask, LAMP generates a complete colonoscopy image while preserving the supplied lesion. (1) \emph{FOV-aware valid-mucosa modeling} separates valid tissue from non-mucosal black regions and restricts contextual reasoning to the legal FOV support, preventing black-region contamination. (2) \emph{Lesion-guided adaptive mucosal context propagation} integrates condition extraction, long-range propagation, and gated fusion. Lesion-to-Mucosa attention extracts lesion appearance cues, and multidirectional Vision Receptance Weighted Key Value (Vision-RWKV)~\cite{duan2024visionrwkv} distributes them across valid mucosa to maintain coherent texture and illumination.

Extensive experiments on five polyp datasets demonstrate state-of-the-art generation quality. Training five representative segmentation models with LAMP-generated pairs further improves overall downstream performance.

Our contributions are fourfold:
\begin{itemize}
    \item To our knowledge, LAMP is the first foreground-guided polyp image synthesis framework, preserving lesions while generating compatible mucosal contexts.

    \item We introduce FOV-aware valid-mucosa modeling to separate the lesion, valid mucosa, and non-mucosal black regions, preventing black-region contamination.

    \item We develop Lesion-Guided Adaptive Mucosal Context Propagation by combining lesion-condition extraction, multidirectional Vision-RWKV propagation, and adaptive residual fusion. It transfers lesion-specific cues across valid mucosa to generate coherent texture.

    \item Extensive experiments show improved generation quality and diversity on five polyp datasets, and better downstream performance across five segmentation models.
\end{itemize}

\section{Related Work}

\minisection{Synthetic Dataset Generation.}
Synthetic imagery has progressed from procedural synthesis, exemplified by early texture generators~\cite{perlin1985image}, to learned generators based on GANs~\cite{goodfellow2020generative}, variational autoencoders~\cite{kingma2014auto,diederik2019introduction}, and diffusion models~\cite{ho2020denoising,song2020denoising}. Synthetic data can improve data-scarce parsing and few-shot detection~\cite{Wang2018SyntheticDM,Lin2023ExploreTP}, but dense prediction additionally requires reliable pixel annotations. DatasetGAN~\cite{zhang2021datasetgan} and BigDatasetGAN~\cite{li2022bigdatasetgan} derive labeled samples from GAN representations. DiffuMask~\cite{wu2023diffumask}, Dataset Diffusion~\cite{nguyen2023dataset}, and DatasetDM~\cite{wu2023datasetdm} extend this direction with diffusion features or annotation-aware generation. These pipelines support scalable synthesis of images with labels but neither preserve the appearance of a supplied lesion nor model endoscopic FOV support. LAMP instead generates a complete image around a concrete lesion and mask condition while explicitly modeling FOV geometry.

\minisection{Polyp Image Generation.}
GAN-based polyp synthesis includes SinGAN-Seg~\cite{thambawita2022singanseg}, bidirectional translation~\cite{qadir2022simple}, and PolypConnect~\cite{fagereng2022polypconnect}. Diffusion methods include mask-conditioned LDM~\cite{machacek2023mask}, Polyp-DDPM~\cite{dorjsembe2024polypddpm}, ControlPolypNet~\cite{sharma2024controlpolypnet}, Polyp-Gen~\cite{liu2025polypgen}, and Polyp-LDM~\cite{qiu2025polypldm}; ControlNet~\cite{zhang2023adding} offers generic spatial conditioning. These methods improve structural control but infer lesion RGB from masks or layouts, potentially weakening correspondence between images and masks. LAMP instead conditions on the lesion foreground and generates its mucosal context.

Foreground-guided generation is most related. LAKE-RED~\cite{zhao2024lakered} retrieves background knowledge from foreground features, FACIG~\cite{chen2025foreground} improves foreground integration and reconstruction, and CamoDreamer~\cite{wang2026camodreamer} separately renders foreground and background before blending. However, they do not propagate lesion-conditioned states across endoscopic mucosa. RWKV~\cite{peng2023rwkv} supports parallel recurrent modeling, and Vision-RWKV~\cite{duan2024visionrwkv} extends it to linear-complexity visual features. LAMP uses Lesion-to-Mucosa attention to extract lesion cues and propagates them in four directions within valid mucosa.

\section{Methodology}

\begin{figure*}
    \centering
    \includegraphics[width=\linewidth]{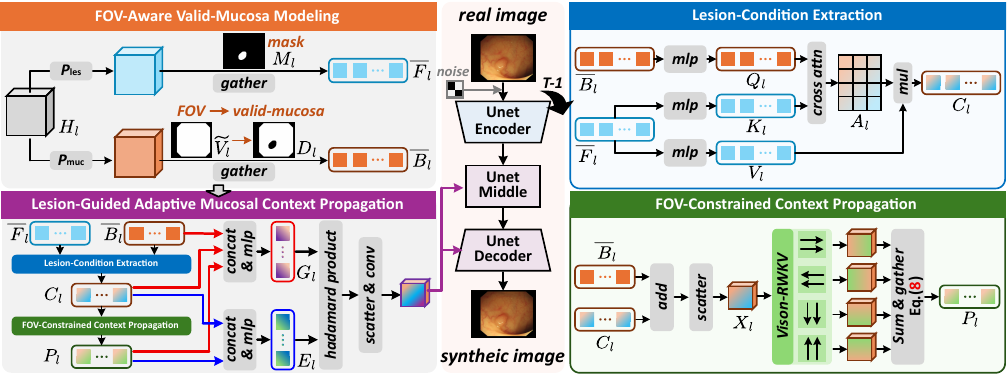}
    \vspace{-8mm}
    \caption{\textbf{Overview of LAMP.}
    (1) FOV-Aware Valid-Mucosa Modeling uses the lesion and FOV masks to partition each feature map into lesion, valid-mucosa, and camera-exterior regions, then gathers separate lesion and mucosal token sequences without exterior interference. (2) Lesion-Guided Adaptive Mucosal Context Propagation transfers lesion appearance to valid mucosa.
    Lesion-Condition Extraction derives location-specific conditions using mucosal queries and lesion keys/values;
    FOV-Constrained Context Propagation spreads them via Vision-RWKV and gathers the outputs.
    }
    \label{fig:model}
    \vspace{-6mm}
\end{figure*}

\subsection{Latent Diffusion Preliminaries}
LAMP builds on a latent diffusion model (LDM)~\cite{rombach2022high}. A frozen encoder maps an image $I\in[0,1]^{3\times H\times W}$ to $z_0=\mathcal E(I)\in\mathbb R^{C_z\times h\times w}$. At diffusion step $t$, Gaussian noise is added as $z_t=\sqrt{\bar\alpha_t}z_0+\sqrt{1-\bar\alpha_t}\epsilon$, where $\epsilon\sim\mathcal N(0,\mathbf I)$. A conditional U-Net $\epsilon_\theta$ predicts the injected noise. Iterative denoising estimates $\widehat z_0$, which a frozen VQ decoder maps back to image space~\cite{esser2021taming}.

\subsection{Problem Formulation and Overview}
Given a real colonoscopy image $I$ and its binary polyp mask $M\in\{0,1\}^{1\times H\times W}$, we construct the foreground condition $F=I\odot M\in[0,1]^{3\times H\times W}$, where $\odot$ denotes the Hadamard (element-wise) product, and $M$ is broadcast across RGB channels and remains the segmentation annotation. An FOV mask $V\in\{0,1\}^{1\times H\times W}$ marks valid endoscopic imaging locations. Each supplied condition must satisfy $M\odot V=M$; conditions extending into the camera exterior are rejected before generation. LAMP models $I_{\mathrm{syn}}=\mathcal G_{\mathrm{LAMP}}(F,M,V;\epsilon)$ with $M_{\mathrm{syn}}=M$.
The foreground may be extracted directly from a real pair or provided by an external augmentation operator.


\Cref{fig:model} summarizes the overall framework.
Following standard LDM, we encode the foreground as $c_F=\mathcal E_{\mathrm{VQ}}(F)$ and resize its mask as $c_M=\operatorname{Resize}(M)$. We then concatenate them to form the spatial condition $c_{\mathrm{sp}}=\operatorname{Cat}(c_F,c_M)$. Starting from Gaussian noise $z_T$, DDIM repeatedly applies the denoising U-Net $\epsilon_\theta$ under $c_{\mathrm{sp}}$ and obtains $\widehat z_0=\operatorname{DDIM}(z_T,c_{\mathrm{sp}},V;\epsilon_\theta)$. Within each selected U-Net block, FOV-aware valid-mucosa modeling first identifies the legal tissue support (\cref{sec:fov_modeling}). Lesion-guided adaptive mucosal context propagation then extracts lesion conditions, propagates them over this support, and writes them back through an adaptive gate (\cref{sec:context_propagation}). Finally, the frozen VQ decoder produces the complete image as $I_{\mathrm{syn}}=\mathcal D_{\mathrm{VQ}}(\widehat z_0)$.

\subsection{FOV-Aware Valid-Mucosa Modeling}
\label{sec:fov_modeling}

\minisection{Endoscopic Imaging Domain.}
Camera-exterior pixels are not tissue and should be excluded from contextual generation.
We separate anatomical content from invalid imaging regions.
We partition the image domain $\Omega$ into anatomical and invalid regions using lesion and FOV masks.
The lesion region $\Omega_{\mathrm{les}}$ contains all positions selected by the polyp mask:
\begin{equation}
\vspace{-2mm}
\Omega_{\mathrm{les}}=\{p\in\Omega\mid M(p)=1\}.
\label{eq:regions}
\vspace{-1mm}
\end{equation}
The valid-mucosa region $\Omega_{\mathrm{muc}}$ contains positions inside the FOV but outside the supplied lesion:
\begin{equation}
\vspace{-2mm}
\Omega_{\mathrm{muc}}=\{p\in\Omega\mid V(p)=1,\ M(p)=0\}.
\label{eq:mucosa_region}
\vspace{-1mm}
\end{equation}
The camera exterior is $\Omega_{\mathrm{out}}=\{p\in\Omega\mid V(p)=0\}$. Thus, $\Omega_V=\Omega_{\mathrm{les}}\mathbin{\dot\cup}\Omega_{\mathrm{muc}}$ is the valid imaging domain and $\Omega=\Omega_V\mathbin{\dot\cup}\Omega_{\mathrm{out}}$. The lesion and mucosa are content roles.

We derive the default $V$ from the original RGB image by thresholding near-black pixels. For the normalized image $I\in[0,1]^{3\times H\times W}$, we use a threshold of $15/255$, equivalent to 15 on the 8-bit intensity scale, giving $Q(p)=\mathbf{1}[\max_{c\in\{R,G,B\}} I_c(p)\leq15/255]$, where $\mathbf{1}[\cdot]$ denotes the indicator function. A candidate region is marked as camera exterior only when it touches the image boundary, giving $V(p)=1-\mathbf{1}[p\in\operatorname{CC}_{\partial}(Q)]$. Thus, an enclosed dark lumen or shadow remains inside the valid FOV. We extract this binary mask once at full resolution and resize it by nearest neighbor for each LAMP block. LAMP can also accept any user-defined binary FOV template satisfying $M\odot V=M$.

\minisection{FOV-Constrained Regional Features.}
Condition extraction requires separate lesion and mucosal representations without camera-exterior interference. We therefore form two legal regional token sequences at each U-Net scale.

Let $H_l\in\mathbb R^{c_l\times h_l\times w_l}$ be the feature map at U-Net scale $l$, where $c_l$, $h_l$, and $w_l$ are its channel number, height, and width. Nearest-neighbor resizing of $M$ and $V$ gives the binary masks $M_l$ and $\widetilde V_l$ at this scale. Their difference defines the valid-mucosa mask $D_l=\widetilde V_l\odot(1-M_l)$. We use two independent learned $1\times1$ projections, $P_{\mathrm{les}}^l:\mathbb R^{c_l}\rightarrow\mathbb R^d$ and $P_{\mathrm{muc}}^l:\mathbb R^{c_l}\rightarrow\mathbb R^d$, to map lesion and mucosal features into a $d=128$ dimensional relation space. For a feature map $X$ and binary mask $A$, $\operatorname{Gather}(X,A)$ scans the spatial locations from left to right and top to bottom, selects the feature vectors where $A=1$, and stacks them as a sequence. We obtain the lesion tokens $\overline F_l$ by gathering the projected features at lesion positions:
\begin{equation}
\vspace{-1.5mm}
\overline F_l=\operatorname{Gather}(P_{\mathrm{les}}^l(H_l),M_l).
\label{eq:regional_features}
\vspace{-0.5mm}
\end{equation}
We obtain the valid-mucosa tokens $\overline B_l$ analogously by gathering the other projection over $D_l$:
\begin{equation}
\vspace{-1.5mm}
\overline B_l=\operatorname{Gather}(P_{\mathrm{muc}}^l(H_l),D_l),
\label{eq:mucosal_features}
\vspace{-0.5mm}
\end{equation}
where $N_f=\sum_p M_l(p)$ and $N_b=\sum_p D_l(p)$ denote the numbers of lesion tokens and valid-mucosa tokens, respectively. Thus, $\overline F_l\in\mathbb R^{N_f\times d}$ contains lesion tokens and $\overline B_l\in\mathbb R^{N_b\times d}$ contains valid-mucosa tokens. The two projections are not shared because the two regions have different roles in the subsequent attention. Camera-exterior locations enter neither token set. We retain the dense masks after gathering to restore the token sequence to its two-dimensional positions and control recurrent state transitions at FOV boundaries.

\subsection{Lesion-Guided Adaptive Mucosal Context Propagation}
\label{sec:context_propagation}

\minisection{Lesion-Condition Extraction.}
The generated mucosa should match the supplied lesion rather than follow an unrelated global condition. We therefore use Lesion-to-Mucosa attention to extract the lesion appearance required at each valid-mucosa location.

The valid-mucosa tokens are projected to queries $Q_l=\overline B_lW_q^l$, while the lesion tokens are projected to keys $K_l=\overline F_lW_k^l$ and values $V_l=\overline F_lW_v^l$. Their multihead aggregation gives the lesion condition $C_l\in\mathbb R^{N_b\times d}$:
\begin{equation}
\vspace{-0.5mm}
C_l=\operatorname{MHA}(Q_l,K_l,V_l).
\label{eq:condition_extraction}
\vspace{-0.5mm}
\end{equation}
For each of four heads, the standard scaled relation $A_{l,r}=\operatorname{softmax}(Q_{l,r}K_{l,r}^{\mathsf T}/\sqrt{d_h})$ aggregates projected lesion values, where $d_h=d/4$ and $A_{l,r}\in\mathbb R^{N_b\times N_f}$. It extracts lesion appearance for every valid-mucosa query.

\minisection{FOV-Constrained Context Propagation.}
Attention conditions each mucosal token independently but does not enforce long-range spatial coherence. We therefore propagate the extracted condition across the valid FOV before fusion.

To recover spatial structure, we scatter the conditioned tokens $\overline B_l+C_l$ to valid-mucosa locations, obtaining $X_l=\operatorname{Scatter}_{D_l}(\overline B_l+C_l)\in\mathbb R^{d\times h_l\times w_l}$. $\operatorname{Scatter}_{A}$ reverses $\operatorname{Gather}$ by restoring tokens where $A=1$ and zeroing all other locations. An FOV-constrained Vision-RWKV~\cite{peng2023rwkv,duan2024visionrwkv} scans $X_l$ from left to right, right to left, top to bottom, and bottom to top, denoted by $\mathcal S=\{\rightarrow,\leftarrow,\downarrow,\uparrow\}$. Horizontal and vertical lines are processed independently, with parameters shared across directions. For direction $s$, $\pi_s$ specifies the scan order and $x_{l,i}^s$ denotes its $i$th feature.

At each position, learned projections produce the receptance $R_{l,i}^s=\sigma(W_r^lx_{l,i}^s)$, key $K_{l,i}^s=W_k^lx_{l,i}^s$, and value $V_{l,i}^s=W_v^lx_{l,i}^s$. Let $z_{l,i-1}^s$ denote the recurrent state accumulated from preceding positions, initialized as $z_{l,0}^s=0$ for each scanned row or column. Here, Vision-RWKV denotes the complete multidirectional visual mixer, while WKV denotes its recurrent weighted key-value operation along one scan direction. The key and value update the state through WKV, while the receptance gates its output:
\begin{equation}
\vspace{-0.5mm}
\begin{aligned}
(\widehat z_{l,i}^s,\widehat q_{l,i}^s)
&=\operatorname{WKV}_l(K_{l,i}^s,V_{l,i}^s,z_{l,i-1}^s),\\
\widehat y_{l,i}^s&=W_o^l(R_{l,i}^s\odot\widehat q_{l,i}^s).
\end{aligned}
\label{eq:rwkv_update}
\vspace{-0.5mm}
\end{equation}
We then constrain the candidate state and output according to the region type:
\begin{equation}
\vspace{-0.5mm}
\left(z_{l,i}^s,y_{l,i}^s\right)=
\begin{cases}
\left(\widehat z_{l,i}^s,\widehat y_{l,i}^s\right), & D_l(\pi_s(i))=1,\\
\left(z_{l,i-1}^s,0\right), & M_l(\pi_s(i))=1,\\
\left(0,0\right), & \widetilde V_l(\pi_s(i))=0.
\end{cases}
\label{eq:fov_rwkv}
\vspace{-0.5mm}
\end{equation}
Specifically, (1) valid mucosa updates the state and emits features; (2) lesion positions carry the state without contributing features; and (3) camera-exterior positions reset it. This allows context to cross the supplied lesion while preventing exterior darkness from entering valid tissue. We restore the directional outputs to the common grid, average them, and gather the valid-mucosa sequence:
\begin{equation}
\vspace{-0.5mm}
P_l=\operatorname{Gather}_{D_l}\!\left(
\frac{1}{|\mathcal S|}\sum_{s\in\mathcal S}
\pi_s^{-1}(y_l^s)\right)
\in\mathbb R^{N_b\times d}.
\label{eq:direction_fusion}
\vspace{-0.5mm}
\end{equation}
This multidirectional recurrence propagates conditioned appearance over long-range mucosa without favoring one image direction, with linear cost in the feature-grid size.

\minisection{Adaptive Residual Fusion.}
Because propagated context is not equally useful at every mucosal location, we learn a spatial gate $G_l$ to control how much conditioned context is written back.
We first transform the extracted and propagated conditions as $E_l=O_l(\operatorname{Cat}(C_l,P_l))\in\mathbb R^{N_b\times d}$. The projection $W_g^l$ maps the concatenated mucosal feature, extracted lesion condition, and propagated state to a scalar gate $G_l$:
\begin{equation}
\vspace{-0.5mm}
G_l=\sigma\!\left(\operatorname{Cat}(\overline B_l,C_l,P_l)W_g^l\right).
\label{eq:gated_residual_fusion}
\vspace{-0.5mm}
\end{equation}
Using this gate, we scatter the transformed context over valid mucosa and write it back residually:
\begin{equation}
\vspace{-0.5mm}
H_l'=H_l+Z_l\!\left(\operatorname{Scatter}_{D_l}(G_l\odot E_l)\right).
\label{eq:residual_writeback}
\vspace{-0.5mm}
\end{equation}
where $G_l\in[0,1]^{N_b\times1}$ and $Z_l$ is a zero-initialized $1\times1$ projection.
Lesion-to-Mucosa attention determines \emph{what} lesion condition is required, context propagation determines \emph{how} it spreads across valid mucosa, the gate determines \emph{how much} is fused, and $D_l$ determines \emph{where} the update is legal.

\subsection{Multiscale Deployment and Learning}
We place LAMP in the U-Net middle and decoder blocks 0/3/6, whose resolutions are $16\times16$, $16\times16$, $32\times32$, and $64\times64$ for a $128\times128$ latent. The blocks use independent projections and share one Vision-RWKV mixer. States are initialized to zero at each denoising step.
We train with real identity pairs. For each $(I,M)$, the foreground $F=I\odot M$ and FOV $V$ condition reconstruction of $I$, providing real co-occurrence between lesions and mucosa.
The objective is the standard $L_1$ noise-prediction loss:
\begin{equation}
\mathcal L_{\mathrm{LAMP}}
=\mathbb E_{z_0,t,\epsilon}\left[
\left\|\epsilon-\epsilon_\theta(z_t,t,c_{\mathrm{sp}};M,V)\right\|_1
\right].
\label{eq:training_loss}
\end{equation}
At inference, $(F,M,V)$ fixes the condition, while initial noise and the DDIM trajectory control mucosal variation. The VQ decoder yields $I_{\mathrm{syn}}=\mathcal D_{\mathrm{VQ}}(\widehat z_0)$ with $M_{\mathrm{syn}}=M$.

\section{Experiments}

\subsection{Experimental Setup}

\minisection{Datasets.}
Experiments are conducted on five datasets: CVC-ClinicDB~\cite{bernal2015wmdova}, Kvasir-SEG~\cite{jha2020kvasirseg}, CVC-ColonDB~\cite{bernal2012automatic}, CVC-300~\cite{vazquez2017benchmark}, and ETIS-LaribPolypDB~\cite{silva2014embedded}. For training, we use 550 image and mask pairs from CVC-ClinicDB and 900 pairs from Kvasir-SEG. For testing, we use 60 pairs from CVC-300, 62 pairs from CVC-ClinicDB, 380 pairs from CVC-ColonDB, 196 pairs from ETIS-LaribPolypDB, and 100 pairs from Kvasir-SEG.

\minisection{Implementation Details.}
LAMP starts from an inpainting LDM~\cite{rombach2022high}. Frozen VQ-f4 maps $512^2$ images to $128^2\times3$ latents. Each context-propagation block uses four-head condition extraction with relation width 128, followed by one Vision-RWKV mixer shared across horizontal and vertical forward and reverse scans. The FOV transition in \cref{eq:fov_rwkv} is applied in every scan. During training, we extract the FOV mask directly from each training image by identifying its boundary-connected near-black regions. At inference, the FOV is an input template and may be defined independently of the source image, provided that it contains the supplied lesion mask. For a simple and controlled comparison, all reported test results use the FOV extracted from the corresponding original image as the template. We train the denoising U-Net and LAMP blocks for 300 epochs with batch size 2 and learning rate $2\times10^{-6}$. All experiments use one NVIDIA Tesla A100 GPU with 40GB memory.

\minisection{Evaluation Metrics.}
We evaluate generative performance using FID~\cite{heusel2017gans}, KID~\cite{binkowski2018demystifying}, and Coverage~\cite{naeem2020reliable}. Lower FID and KID indicate better distributional fidelity, while higher Coverage indicates that the generated distribution covers more of the real-image distribution.

\begin{table*}[t]
    \footnotesize
    \centering
    \renewcommand{\arraystretch}{1.0}
    \setlength\tabcolsep{2.2pt}
    \caption{\textbf{Quantitative results.} Lower FID and KID and higher Coverage indicate better generative performance. Background-guided methods use polyp-free SUN backgrounds; all methods receive matched foreground conditions.}
    \vspace{-2mm}
    \label{tab:method_compared}
    \resizebox{\textwidth}{!}{%
    \begin{tabular}{ll*{18}{c}}
    \toprule
    \multirow{2}{*}{Methods} & \multirow{2}{*}{Input}
    & \multicolumn{3}{c}{\textbf{CVC-300}}
    & \multicolumn{3}{c}{\textbf{CVC-ClinicDB}}
    & \multicolumn{3}{c}{\textbf{CVC-ColonDB}}
    & \multicolumn{3}{c}{\textbf{ETIS-Larib.}}
    & \multicolumn{3}{c}{\textbf{Kvasir-SEG}}
    & \multicolumn{3}{c}{\textbf{Overall}}\\
    \cmidrule(lr){3-5}\cmidrule(lr){6-8}\cmidrule(lr){9-11}
    \cmidrule(lr){12-14}\cmidrule(lr){15-17}\cmidrule(lr){18-20}
    &&FID$\downarrow$&KID$\downarrow$&Cov.$\uparrow$
    &FID$\downarrow$&KID$\downarrow$&Cov.$\uparrow$
    &FID$\downarrow$&KID$\downarrow$&Cov.$\uparrow$
    &FID$\downarrow$&KID$\downarrow$&Cov.$\uparrow$
    &FID$\downarrow$&KID$\downarrow$&Cov.$\uparrow$
    &FID$\downarrow$&KID$\downarrow$&Cov.$\uparrow$\\
    \midrule
    \rowcolor{gray!15}AdaIN~\cite{huang2017arbitrary}$_{17}$&$\mathcal F+\mathcal B$&221.54&0.137&0.067&201.27&0.090&0.339&152.61&0.105&0.050&147.25&0.085&0.041&160.53&0.067&0.460&104.75&0.052&0.160\\
    DCI~\cite{zhang2020deep}$_{20}$&$\mathcal F+\mathcal B$&223.31&0.135&0.083&206.12&0.097&0.403&148.01&0.095&0.068&146.64&0.081&0.046&181.77&0.087&0.290&105.05&0.046&0.149\\
    \rowcolor{gray!15}LCGNet~\cite{li2023location}$_{23}$&$\mathcal F+\mathcal B$&227.40&0.146&0.067&198.18&0.091&0.306&153.30&0.100&0.047&144.28&0.082&0.036&183.25&0.095&0.410&110.84&0.052&0.160\\
    \midrule
    \rowcolor{gray!15}TFill~\cite{zheng2022bridging}$_{22}$&$\mathcal F$&311.85&0.278&0.017&236.07&0.107&0.242&217.34&0.174&0.018&241.84&0.216&0.010&201.95&0.120&0.310&173.31&0.136&0.066\\
    RePaint-L~\cite{lugmayr2022repaint}$_{22}$&$\mathcal F$&190.72&0.109&0.083&225.53&0.137&0.355&125.26&0.072&0.108&176.21&0.130&0.071&161.19&0.088&0.510&102.05&0.058&0.169\\
    \rowcolor{gray!15}LAKE-RED~\cite{zhao2024lakered}$_{24}$&$\mathcal F$&187.85&0.086&0.033&162.06&0.036&0.742&106.54&0.042&0.129&224.32&0.173&0.046&136.78&0.041&0.730&93.72&0.044&0.232\\
    FACIG~\cite{chen2025foreground}$_{25}$&$\mathcal F$&230.19&0.122&0.000&185.79&0.052&0.532&134.14&0.064&0.039&228.08&0.165&0.041&166.54&0.066&0.450&108.35&0.052&0.133\\
    \rowcolor{gray!15}CamoDreamer~\cite{wang2026camodreamer}$_{26}$&$\mathcal F$&146.27&0.052&0.083&152.31&0.032&0.806&89.95&0.027&0.213&174.59&0.113&0.061&131.95&0.041&0.820&74.35&0.025&0.301\\
    \rowcolor{c1!30}\textbf{LAMP (Ours)}&$\mathcal F$&\textbf{139.33}&\textbf{0.050}&\textbf{0.150}&\textbf{146.96}&0.036&\textbf{0.919}&\textbf{78.75}&\textbf{0.024}&\textbf{0.329}&\textbf{114.18}&\textbf{0.057}&\textbf{0.168}&\textbf{114.86}&\textbf{0.027}&\textbf{0.970}&\textbf{61.02}&\textbf{0.018}&\textbf{0.397}\\
    \bottomrule
    \end{tabular}}
    \vspace{-1.2mm}
\end{table*}

\begin{figure*}[!t]
    \centering
    \includegraphics[width=\textwidth]{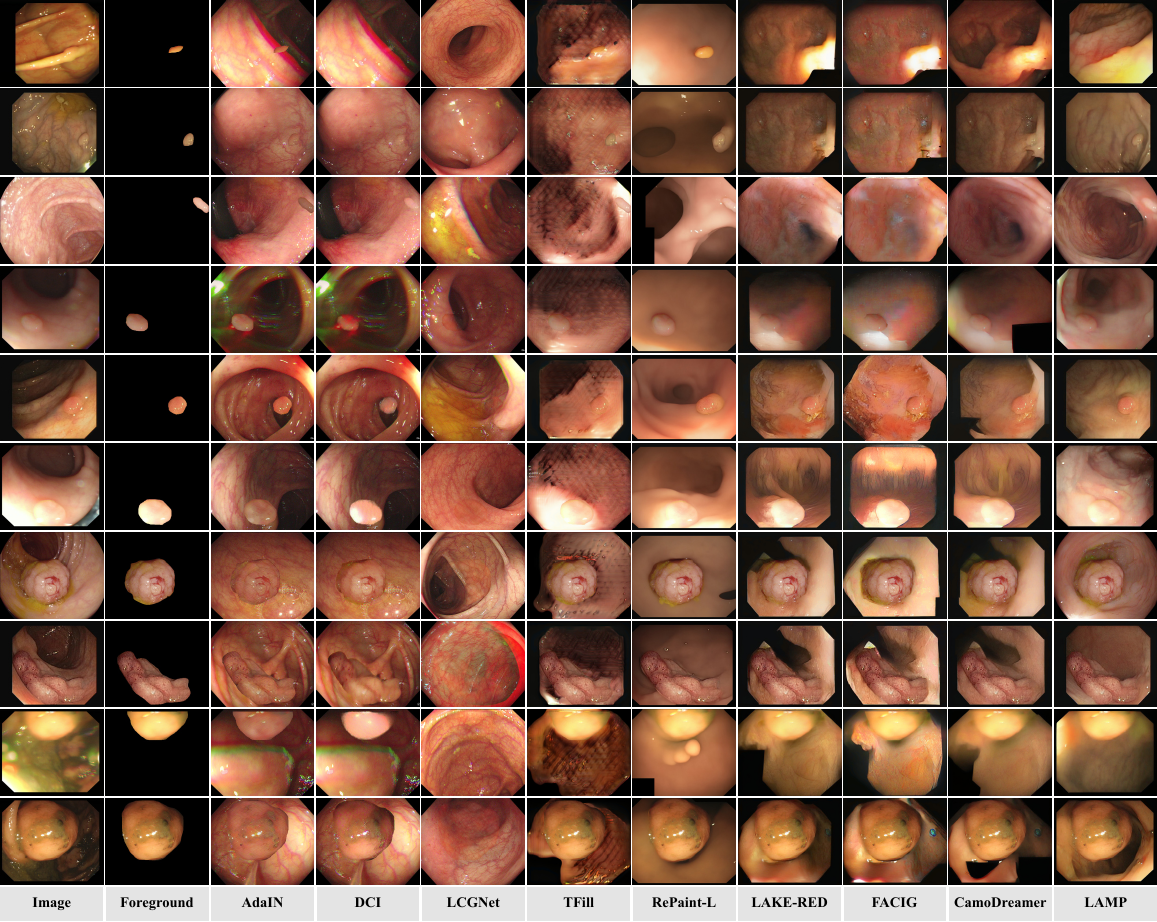}
    \vspace{-6.5mm}
    \caption{\textbf{Qualitative comparison.} Columns 1 and 2 show the original image and foreground.
    Columns 3 to 11 show AdaIN~\cite{huang2017arbitrary}, DCI~\cite{zhang2020deep}, LCGNet~\cite{li2023location}, TFill~\cite{zheng2022bridging}, RePaint-L~\cite{lugmayr2022repaint}, LAKE-RED~\cite{zhao2024lakered}, FACIG~\cite{chen2025foreground}, CamoDreamer~\cite{wang2026camodreamer} and our LAMP, which generates cleaner and more coherent mucosa.}
    \label{fig:qualitative_analysis}
    \vspace{-6mm}
\end{figure*}

\minisection{Comparison Protocol.}
We compare LAMP with eight representative methods. Background-guided AdaIN~\cite{huang2017arbitrary}, DCI~\cite{zhang2020deep}, and LCGNet~\cite{li2023location} receive a foreground $\mathcal F$ and target background $\mathcal B$ drawn from a dedicated polyp-free SUN subset~\cite{misawa2021development}. Foreground-guided TFill~\cite{zheng2022bridging}, RePaint-L~\cite{lugmayr2022repaint}, LAKE-RED~\cite{zhao2024lakered}, FACIG~\cite{chen2025foreground}, and CamoDreamer~\cite{wang2026camodreamer}\footnote{Since official implementations of FACIG and CamoDreamer are unavailable, we reimplement both methods according to their published descriptions.} receive the same $\mathcal F$ without a selected background; LAMP additionally uses its FOV. For a fair comparison, all competing methods are trained on the same polyp training set, with hyperparameters following their original papers.

\begin{table*}[t]
\begin{minipage}[t]{0.49\textwidth}
    \centering
    \caption{\textbf{Ablation of FOV-aware modeling and lesion-guided adaptive mucosal context propagation.}}
    \vspace{-2mm}
    \label{tab:component_ablation}
    \setlength{\tabcolsep}{12pt}
    \renewcommand{\arraystretch}{0.85}
    \resizebox{\linewidth}{!}{%
    \begin{tabular}{l|ccc}
    \toprule
    Setting&FID$\downarrow$&KID$\downarrow$&Coverage$\uparrow$\\
    \midrule
    \circlednum{1} w/o FOV Modeling& 77.81&0.031&0.291\\
    \circlednum{2} w/o Condition Extraction&72.83&0.027&0.311\\
    \circlednum{3} w/o Context Propagation&74.25&0.029&0.308\\
    \circlednum{4} w/o Adaptive Gate&67.31&0.021&0.370\\
    \rowcolor{c1!30}\circlednum{5} \textbf{LAMP (Ours)}&\textbf{61.02}&\textbf{0.018}&\textbf{0.397}\\
    \bottomrule
    \end{tabular}}
\end{minipage}\hfill%
\begin{minipage}[t]{0.49\textwidth}
    \centering
    \caption{\textbf{Ablation of LAMP deployment across U-Net stages.}}
    \vspace{-2mm}
    \label{tab:injection_ablation}
    \setlength{\tabcolsep}{16pt}
    \setlength{\extrarowheight}{3.35pt}
    \resizebox{\linewidth}{!}{%
    \begin{tabular}{l|ccc}
    \toprule
    Position&FID$\downarrow$&KID$\downarrow$&Coverage$\uparrow$\\
    \midrule
    \circlednum{1} Encoder&65.71&0.020&0.371\\
    \circlednum{2} Middle&70.32&0.025&0.336\\
    \circlednum{3} Decoder&62.53&0.019&0.379\\
    \rowcolor{c1!30}\circlednum{4} \makecell{\textbf{Middle+Decoder}\\\textbf{(Ours)}}&\textbf{61.02}&\textbf{0.018}&\textbf{0.397}\\
    \bottomrule
\end{tabular}}
\end{minipage}
\end{table*}

\subsection{Polyp Image Generation Results}

\begin{table*}[t]
    \footnotesize
    \centering
    \renewcommand{\arraystretch}{0.85}
    \setlength\tabcolsep{7pt}
    \caption{\textbf{Downstream segmentation performance.} ``w/o Syn.'' uses real training data only, while ``w/ Syn.'' adds 1,000 image and mask pairs generated by LAMP. Higher $S_\alpha$, $E_{\phi}^{mn}$, $F_\beta^w$, Dice, and IoU and lower MAE indicate better performance.}
    \vspace{-1mm}
    \resizebox{\textwidth}{!}{%
    \begin{tabular}{cc*{5}{c>{\columncolor{c1!30}}c}}
    \toprule
    \multirow{2}{*}{Dataset} & \multirow{2}{*}{Metric}
    & \multicolumn{2}{c}{PraNet~\cite{fan2020pranet}$_{20}$}
    & \multicolumn{2}{c}{SEPnet~\cite{wang2024sepnet}$_{24}$}
    & \multicolumn{2}{c}{STDDNet~\cite{chen2025stddnet}$_{25}$}
    & \multicolumn{2}{c}{CMSA~\cite{wang2026cmsanet}$_{26}$}
    & \multicolumn{2}{c}{Artemis~\cite{wang2026artemis}$_{26}$} \\
    \cmidrule(lr){3-4} \cmidrule(lr){5-6} \cmidrule(lr){7-8}
    \cmidrule(lr){9-10} \cmidrule(lr){11-12}
    && w/o Syn. & w/ Syn. & w/o Syn. & w/ Syn. & w/o Syn. & w/ Syn.
    & w/o Syn. & w/ Syn. & w/o Syn. & w/ Syn. \\
    \midrule
        \multirow{6}{*}{CVC-300}
    & $S_\alpha\uparrow$ & 0.925 & 0.931 & 0.944 & 0.946 & 0.930 & 0.945 & 0.931 & 0.946 & 0.942 & 0.947 \\
    & $E_{\phi}^{mn}\uparrow$ & 0.946 & 0.950 & 0.965 & 0.968 & 0.943 & 0.965 & 0.965 & 0.965 & 0.965 & 0.968 \\
    & $F_\beta^w\uparrow$ & 0.843 & 0.856 & 0.883 & 0.883 & 0.852 & 0.883 & 0.882 & 0.885 & 0.884 & 0.889 \\
    & Dice$\uparrow$ & 0.871 & 0.877 & 0.899 & 0.905 & 0.884 & 0.895 & 0.897 & 0.897 & 0.896 & 0.909 \\
    & IoU$\uparrow$ & 0.797 & 0.811 & 0.831 & 0.841 & 0.815 & 0.828 & 0.828 & 0.832 & 0.831 & 0.846 \\
    & MAE$\downarrow$ & 0.010 & 0.009 & 0.006 & 0.006 & 0.009 & 0.006 & 0.008 & 0.006 & 0.006 & 0.006 \\
    \midrule
    \multirow{6}{*}{CVC-ClinicDB}
    & $S_\alpha\uparrow$ & 0.936 & 0.942 & 0.959 & 0.960 & 0.944 & 0.960 & 0.949 & 0.960 & 0.944 & 0.960 \\
    & $E_{\phi}^{mn}\uparrow$ & 0.963 & 0.965 & 0.982 & 0.984 & 0.971 & 0.981 & 0.981 & 0.987 & 0.977 & 0.981 \\
    & $F_\beta^w\uparrow$ & 0.896 & 0.900 & 0.932 & 0.933 & 0.914 & 0.932 & 0.932 & 0.941 & 0.927 & 0.931 \\
    & Dice$\uparrow$ & 0.899 & 0.906 & 0.933 & 0.937 & 0.920 & 0.932 & 0.933 & 0.940 & 0.932 & 0.933 \\
    & IoU$\uparrow$ & 0.849 & 0.858 & 0.885 & 0.887 & 0.875 & 0.883 & 0.886 & 0.893 & 0.889 & 0.884 \\
    & MAE$\downarrow$ & 0.009 & 0.008 & 0.007 & 0.006 & 0.010 & 0.007 & 0.007 & 0.007 & 0.008 & 0.007 \\
    \midrule
    \multirow{6}{*}{CVC-ColonDB}
    & $S_\alpha\uparrow$ & 0.820 & 0.859 & 0.878 & 0.892 & 0.875 & 0.880 & 0.861 & 0.877 & 0.875 & 0.895 \\
    & $E_{\phi}^{mn}\uparrow$ & 0.847 & 0.886 & 0.914 & 0.922 & 0.909 & 0.911 & 0.906 & 0.912 & 0.915 & 0.926 \\
    & $F_\beta^w\uparrow$ & 0.699 & 0.757 & 0.799 & 0.817 & 0.791 & 0.797 & 0.794 & 0.810 & 0.804 & 0.822 \\
    & Dice$\uparrow$ & 0.712 & 0.777 & 0.819 & 0.834 & 0.814 & 0.816 & 0.810 & 0.811 & 0.824 & 0.839 \\
    & IoU$\uparrow$ & 0.640 & 0.706 & 0.740 & 0.761 & 0.735 & 0.739 & 0.731 & 0.733 & 0.748 & 0.767 \\
    & MAE$\downarrow$ & 0.043 & 0.034 & 0.028 & 0.024 & 0.030 & 0.027 & 0.030 & 0.028 & 0.025 & 0.022 \\
    \midrule
    \multirow{6}{*}{ETIS-LaribPolypDB}
    & $S_\alpha\uparrow$ & 0.794 & 0.823 & 0.881 & 0.894 & 0.859 & 0.895 & 0.873 & 0.895 & 0.877 & 0.892 \\
    & $E_{\phi}^{mn}\uparrow$ & 0.808 & 0.856 & 0.899 & 0.910 & 0.842 & 0.914 & 0.899 & 0.910 & 0.907 & 0.908 \\
    & $F_\beta^w\uparrow$ & 0.600 & 0.657 & 0.754 & 0.774 & 0.716 & 0.773 & 0.770 & 0.777 & 0.763 & 0.771 \\
    & Dice$\uparrow$ & 0.628 & 0.686 & 0.795 & 0.811 & 0.765 & 0.810 & 0.795 & 0.813 & 0.804 & 0.808 \\
    & IoU$\uparrow$ & 0.567 & 0.619 & 0.718 & 0.735 & 0.698 & 0.732 & 0.718 & 0.735 & 0.728 & 0.732 \\
    & MAE$\downarrow$ & 0.031 & 0.021 & 0.016 & 0.014 & 0.025 & 0.014 & 0.015 & 0.012 & 0.014 & 0.012 \\
    \midrule
    \multirow{6}{*}{Kvasir-SEG}
    & $S_\alpha\uparrow$ & 0.915 & 0.917 & 0.931 & 0.934 & 0.926 & 0.929 & 0.926 & 0.927 & 0.925 & 0.935 \\
    & $E_{\phi}^{mn}\uparrow$ & 0.941 & 0.944 & 0.956 & 0.961 & 0.949 & 0.954 & 0.951 & 0.961 & 0.957 & 0.962 \\
    & $F_\beta^w\uparrow$ & 0.885 & 0.890 & 0.909 & 0.914 & 0.895 & 0.906 & 0.904 & 0.916 & 0.910 & 0.910 \\
    & Dice$\uparrow$ & 0.898 & 0.900 & 0.911 & 0.922 & 0.915 & 0.912 & 0.909 & 0.919 & 0.916 & 0.921 \\
    & IoU$\uparrow$ & 0.840 & 0.846 & 0.861 & 0.869 & 0.855 & 0.866 & 0.854 & 0.868 & 0.863 & 0.870 \\
    & MAE$\downarrow$ & 0.030 & 0.028 & 0.022 & 0.021 & 0.028 & 0.024 & 0.026 & 0.022 & 0.023 & 0.021 \\
    \bottomrule
    \end{tabular}}
    \label{tab:downstream_segmentation}
    \vspace{-2mm}
\end{table*}

\minisection{Quantitative Results.}
\Cref{tab:method_compared} shows that LAMP achieves the strongest overall generation performance. Against CamoDreamer, it reduces overall FID from 74.35 to 61.02 (17.9\%) and KID from 0.025 to 0.018 (28.0\%), while increasing Coverage from 0.301 to 0.397 by 9.6 percentage points (31.9\% relatively). These gains validate contamination-free, lesion-guided mucosal generation.

\minisection{Qualitative Results.}
\Cref{fig:qualitative_analysis} groups small, medium, and large lesions in rows 1--3, 4--6, and 7--10. Baselines exhibit photometric seams, weakened boundaries, or globally inconsistent mucosa. LAMP preserves the supplied lesion with smoother transitions, coherent folds and vessels, and cleaner FOV boundaries. Lesion-to-Mucosa attention matches appearance, Vision-RWKV propagates it across mucosa, and FOV-constrained transitions block camera-exterior leakage.

\minisection{Ablation Study.}
\Cref{tab:component_ablation} isolates each component. Setting \circlednum{1} replaces $D_l$ with $1-M_l$ and disables FOV resets, causing camera-exterior pixels to enter mucosal reasoning. Restoring FOV modeling reduces FID/KID by 21.6\%/41.9\% and raises Coverage by 10.6 points, the largest change, confirming its role in preventing exterior contamination. Setting \circlednum{2} removes Lesion-to-Mucosa attention, leaving Vision-RWKV without explicit lesion cues. Its restoration yields 16.2\%/33.3\% lower FID/KID and 8.6-point higher Coverage, showing that lesion-dependent context is essential. Setting \circlednum{3} bypasses Vision-RWKV and writes attention output directly through the gate. Restoring propagation improves FID/KID by 17.8\%/37.9\% and Coverage by 8.9 points, showing that local attention alone lacks long-range coherence. Setting \circlednum{4} fixes $G_l=1$ for uniform fusion; adaptive gating improves FID/KID by 9.3\%/14.3\% and Coverage by 2.7 points through spatially selective updates.

\Cref{tab:injection_ablation} compares deployment positions. Encoder-only features are diluted by later blocks, while middle-only deployment captures global structure but lacks multiscale refinement. Decoder-only is thus the best single-stage setting. Adding the middle block further reduces FID/KID by 2.4\%/5.3\% and raises Coverage by 1.8 points. The middle block establishes global compatibility between the lesion and mucosa at low resolution, while decoder scales progressively recover folds, vessels, and illumination details, explaining the optimal Middle+Decoder setting.

\subsection{Downstream Application}
LAMP focuses on generating compatible mucosa around a supplied foreground; it does not claim foreground deformation as a method contribution. To evaluate data augmentation with lesion-level variation, we adopt a variational Atlas-SVF model~\cite{dalca2019unsupervised}, which samples stationary velocity fields and integrates them into smooth diffeomorphic transformations for aligned deformation of the foreground and mask.

\begin{figure}[t]
    \centering
    \includegraphics[width=\linewidth]{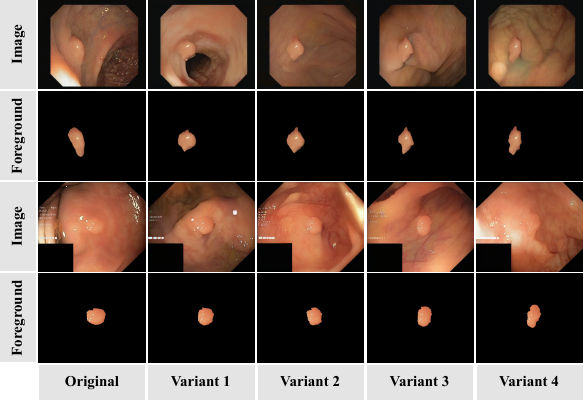}
    \vspace{-6mm}
    \caption{\textbf{LAMP combines diverse lesion deformation with compatible mucosal synthesis.} Column 1 shows the original image. Columns 2 to 5, labeled Variants 1 to 4, show synthetic images obtained by deforming the foreground and mask with Atlas-SVF and generating compatible mucosa.
    }
    \label{fig:diverse_foreground}
    \vspace{-2mm}
\end{figure}

As illustrated in \cref{fig:diverse_foreground}, column 1 shows the original image, while columns 2 to 5 show four complete synthetic variants obtained from independently sampled deformations.
For each variant, the diffeomorphic transformation is applied to both foreground RGB and its binary mask through the same differentiable spatial sampler~\cite{jaderberg2015spatial}. The resulting deformed foreground and aligned mask are then provided to LAMP, which generates the corresponding FOV-consistent mucosa. The results exhibit coherent mucosal textures with no contamination from black camera borders.

\begin{figure}[t]
    \centering
    \includegraphics[width=\linewidth]{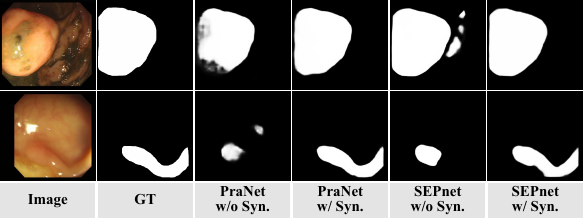}
    \vspace{-6mm}
    \caption{\textbf{LAMP-generated augmentation improves polyp segmentation.} Predictions under ``w/o Syn.'' and ``w/ Syn.'' settings are shown for small, low-contrast, and irregular lesions.}
    \label{fig:prediction_result}
    \vspace{-6mm}
\end{figure}

\minisection{Downstream Protocol.}
We add exactly 1,000 synthetic image and mask pairs generated by LAMP to the real training set.
Under the same augmentation budget, we train PraNet~\cite{fan2020pranet}, SEPnet~\cite{wang2024sepnet}, STDDNet~\cite{chen2025stddnet}, CMSA~\cite{wang2026cmsanet}, and Artemis~\cite{wang2026artemis}. ``w/o Syn.'' uses only real data and ``w/ Syn.'' adds the same synthetic set. Five benchmarks are evaluated with $S_\alpha$~\cite{fan2017structure}, $E_{\phi}^{mn}$~\cite{fan2018enhanced}, $F_\beta^w$~\cite{margolin2014evaluate}, Dice, IoU, and MAE.

\minisection{Downstream Results.}
To further assess the utility of LAMP-generated images for downstream perception, we augment the real training set with 1,000 synthetic image and mask pairs and train five representative polyp segmentation models. Quantitative and qualitative results are reported in \cref{tab:downstream_segmentation,fig:prediction_result}. Overall, LAMP augmentation improves segmentation performance across the evaluated architectures and test domains. On average, Dice and IoU increase by 1.32 and 1.44 percentage points, respectively, while MAE decreases by 15.9\%. The improvements are more pronounced on the challenging CVC-ColonDB and ETIS datasets, where average Dice rises by 1.96 and 2.82 points. In particular, PraNet gains 6.5 Dice and 6.6 IoU points on CVC-ColonDB, while STDDNet gains 4.5 Dice points and reduces MAE by 44.0\% on ETIS. These results indicate that preserving exact alignment between the foreground and mask while generating diverse, lesion-compatible mucosal contexts reduces the domain gap to real colonoscopy images without introducing label noise, thereby improving overall downstream segmentation performance.

\section{Conclusion}
In this paper, we propose LAMP, a novel framework for FOV-aware foreground-guided polyp image synthesis. LAMP explicitly distinguishes the supplied lesion, valid mucosa, and camera exterior to constrain contextual reasoning within anatomically legal tissue regions, thereby preventing exterior darkness from contaminating generated mucosa. In addition, it constructs lesion-compatible mucosal context by combining Lesion-to-Mucosa attention with FOV-constrained multidirectional Vision-RWKV propagation and adaptive residual fusion, which jointly improve local appearance compatibility and long-range texture coherence. Extensive quantitative and qualitative experiments on five polyp datasets validate the effectiveness of the proposed framework, achieving state-of-the-art generation quality and diversity, while augmentation experiments across five segmentation models demonstrate its strong potential for improving downstream polyp segmentation.

\bibliographystyle{IEEEtran}
\bibliography{egbib}

\end{document}